\documentclass{article}

\usepackage{arxiv}

\usepackage[utf8]{inputenc} 
\usepackage[T1]{fontenc}    
\usepackage{hyperref}       
\usepackage{url}            
\usepackage{booktabs}       
\usepackage{amsfonts}       
\usepackage{nicefrac}       
\usepackage{microtype}      
\usepackage{cleveref}       
\usepackage{lipsum}         
\usepackage{graphicx}
\usepackage[round]{natbib}
\usepackage{doi}

\pdfoutput=1

\usepackage{hyperref}

\hypersetup{
    colorlinks=true,
    linkcolor=black,
    citecolor=black,
    filecolor=black,
    urlcolor=black,
}

\title{Fast Learning of Dynamic Hand Gesture Recognition with Few-Shot Learning Models}

\date{December 16, 2022}

\author{ \href{https://orcid.org/0000-0001-7851-3015}{\includegraphics[scale=0.06]{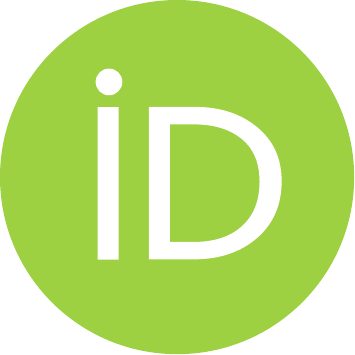}\hspace{1mm}Niels Schlüsener} \\
	Münster School of Business\\
	FH Münster - University of Applied Sciences\\
	Münster, 48149 \\
	\texttt{niels.schluesener@fh-muenster.de} \\
	\And
	\href{https://orcid.org/0000-0003-0045-8460}{\includegraphics[scale=0.06]{arxiv-orcid.pdf}\hspace{1mm}Michael Bücker} \\
	Münster School of Business\\
	FH Münster - University of Applied Sciences\\
	Münster, 48149 \\
	\texttt{michael.buecker@fh-muenster.de} \\
}

\renewcommand{\headeright}{}
\renewcommand{\undertitle}{} 
\renewcommand{\shorttitle}{Fast Learning of Dynamic Hand Gesture Recognition with Few-Shot Learning Models}

\hypersetup{
pdftitle={Fast Learning of Dynamic Hand Gesture Recognition with Few-Shot Learning Models},
pdfsubject={q-bio.NC, q-bio.QM},
pdfauthor={Niels Schlüsener, Michael Bücker},
pdfkeywords={First keyword, Second keyword, More},
}

\begin{document}

\maketitle

\begin{abstract}We develop Few-Shot Learning models trained to recognize five or ten different dynamic hand gestures, respectively, which are arbitrarily interchangeable by providing the model with one, two, or five examples per hand gesture. All models were built in the Few-Shot Learning architecture of the Relation Network (RN), in which Long-Short-Term Memory cells form the backbone. The models use hand reference points extracted from RGB-video sequences of the Jester dataset which was modified to contain 190 different types of hand gestures. Result show accuracy of up to 88.8\% for recognition of five and up to 81.2\% for ten dynamic hand gestures. The research also sheds light on the potential effort savings of using a Few-Shot Learning approach instead of a traditional Deep Learning approach to detect dynamic hand gestures. Savings were defined as the number of additional observations required when a Deep Learning model is trained on new hand gestures instead of a Few Shot Learning model. The difference with respect to the total number of observations required to achieve approximately the same accuracy indicates potential savings of up to 630 observations for five and up to 1260 observations for ten hand gestures to be recognized. Since labeling video recordings of hand gestures implies significant effort, these savings can be considered substantial.
\end{abstract}


\section{Introduction}\label{sec1}

Dynamic hand gesture recognition (DHGR) is a widely researched field of applied machine learning with use cases in robotics \citep{yangEtal2017}, the medical sector \citep{galloEtal2011}, gaming \citep{leeHong2010}, and sign language translation \citep{adeyanjuEtal2021}. However, the majority of previous research only extends to the recognition of a fixed set of gestures with standard machine learning (SML) approaches \citep{yasenJusoh2019}. To ensure highly accurate models, such approaches require a lot of data for each gesture and need to be retrained every time  gestures are adapted, added, or exchanged, which in practice is costly and time-consuming. Applications in which the hand gestures are frequently changed or individually defined by the user are limited by these factors. Examples include an individually configurable robotic arm adapted for the requirements of a single user or a sandbox game that is controlled by hand gestures designed by the player. To overcome these restrictions, some researchers apply Few Shot Learning (FSL) or Meta Learning approaches to the task of DHGR \citep{hazraEtAl2019, fanEtAl2020, zengEtal2021, mauroEtal2022, shenEtal2022, rzeckiKrzysztof2020, rahimianEtal2021, zabihiEtal2022, luiMan2012, wanEtal2013, jiangEtal2013, cabreraEtal2017, jiangEtal2017, zhangEtal2017, chandrasekharMhala2019, luEtal2019, maEtal2020, wuEtal2021}. These methods that generalize their learnings aiming to adapt to new classes with just a few examples per class, have been extensively studied with great success for the area of object recognition \citep{wangEtal2020}, but have received less attention in the field of DHGR. The few existing studies can be grouped by the type of data they use to recognize the gestures:  models that operate with (a) radar waves \citep{hazraEtAl2019, fanEtAl2020, zengEtal2021, mauroEtal2022, shenEtal2022}, (b) electromyography \citep{rzeckiKrzysztof2020, rahimianEtal2021, zabihiEtal2022} or (c) image data\citep{luiMan2012, wanEtal2013, jiangEtal2013, cabreraEtal2017, jiangEtal2017, zhangEtal2017, chandrasekharMhala2019, luEtal2019, maEtal2020, wuEtal2021}. Among the publications that used image data, the vast majority of models were trained with RGB-D data, which contains three RGB channels and their corresponding depth information. Despite the excellent results of these approaches, the fact that RGB-D cameras are nowhere near as common as RGB cameras creates a barrier to applying these models in practice. In order to overcome this hurdle, this study proposes FSL models for DHGR that operate solely on the basis of RGB data. In addition, we investigate the value of applying FSL to the task of DHGR by measuring the reduced number of samples using our FSL model compared to the number of samples required using an SML approach with comparable accuracy. Hence, this work quantifies a frequently cited advantage of FSL over SML mentioned in many publications \citep{luEtal2019, maEtal2020, wuEtal2021}.

The contributions of this paper are as follows:

\begin{itemize}
 \item Six embedding-based FSL models, trained to quickly adapt to the task of recognizing five or ten dynamic hand gestures by providing  one, two, or five RGB video samples per gesture. 
 \item Comprehensive experimental results on a modified dataset and a benchmark dataset.
 \item Quantification of the FSL models' benefits compared to the use of SML models.
\end{itemize}

\section{Related Work}\label{sec2}
\subsection{Dynamic Hand Gesture Recognition}\label{sec2}

Many approaches follow a similar process for machine-aided recognition of dynamic hand gestures. Starting by recording a hand gesture performed by a human to capture its data, this information then gets processed to convert it to low dimensional features, which are fed into an algorithm that recognizes the gesture \citep{pisharadyEtal2015}. Reviews on DHGR systems show, that previous research pursued different approaches for each of the three-phase steps, data collection, feature extraction, and classification \citep{pisharadyEtal2015, yasenJusoh2019, oudahEtal2020, norainiEtal2021}. Early studies on DHGR focused on RGB cameras or data gloves to capture data, created features manually, and used algorithms like Dynamic Time Warping \citep{berndtClifford1994} or Hidden Markov Models \citep{rabinderJuang1986} for classification. In recent works, bulky data gloves have been exchanged for smaller devices or touchless sensors. As for image data, the development of depth sensors, such as Microsoft Kinect or the Leap Motion controller, has led researchers to focus on RGB-D data since it offers rich 3D information, which can easily be processed into precise skeleton data \citep{zhang2012}. It is worth mentioning that skeleton data, used frequently in studies where raw data is processed into effective features, can also be extracted from RGB images thanks to the latest research with libraries like Google's MediaPipe \citep{zhangEtal2020}. In the area of recognition algorithms, deep learning methods such as Convolutional Neural Networks (CNN) \citep{lecunEtal2015} or Long Short-Term Memory Networks (LSTM) \citep{hochreiterSchmidhuber1997} have supplemented the rather traditional approaches. Despite the high accuracy of these algorithms, the issue of adapting to new, previously unknown gestures still exists. The actual use of these techniques in real applications is constrained by the necessity of a significant amount of training data and the need for the model to be retrained after acquiring new data.

\subsection{Few Shot Learning}\label{sec2}

Few Shot Learning is an approach to solve the limitations of deep learning algorithms when frequently adapting to new tasks. Following the ML definition of \cite{mitchell1997}, \cite{wangEtal2020} describe FSL as a type of machine learning problem (specified by E and T, where E contains only a limited number of examples with supervised information for the target T. Recent research on FSL problems mainly studies supervised learning tasks in image classification \citep{liuEtal2018, dhillonEtal2019}, text classification \citep{yanEtal2018} or object recognition \citep{kangEtal2019, fanEtal2020b}. 

FSL tasks are defined by N-Way, the number of exchangeable classes to be classified, and K-Shot, the number of new samples required for each class. These samples are stored in the support set, while the ones to be predicted are called the query set. If K-Shot equals one, we refer to One Shot Learning, while approaches that recognize unseen classes without them appearing in the support set are Zero Shot Learning models. Initially, FSL models are meta-trained on a wide variety of tasks, drawn in iterations from a huge dataset, with the goal of teaching the model the ability to adapt quickly to new tasks, given only a few samples per iteration \citep{chenEtal2019}. A high level of task variety strengthens the learning success in this process. Hence datasets with a large variety of classes are used frequently for meta-learning, e.g. Omniglot \citep{lakeEtal2019} or ImageNet \citep{dengEtal2009}. 

Various methods are applied in FSL research, which may be categorized as data, model, and algorithm approaches according to the taxonomy of \cite{wangEtal2020}. Data methods use prior knowledge to augment the support set in order to increase the number of samples up until the point where SML algorithms can be trained with. Algorithm methods are trained to quickly search for the best parameters to solve a new task by either providing a good initialization or guiding the search steps efficiently. Model methods are able to constrain the complexity of new tasks, e.g. by turning samples into highly embedded features, easily distinguishable from each other. Among the so-called embedding learners, approaches such as Matching Nets \citep{vinyalsEtal2016}, Prototypical Networks \citep{snellEtal2017}, and Relation Networks \citep{sungEtal2018} are frequently used in research.

\subsection{Dynamic Hand Gesture Recognition with Few Shot Learning}\label{sec2}

Intending to overcome the limitations of SML models, various researchers in the field of DHGR apply FSL to their problems. Among 18 papers analyzed from 2012 to 2022, trends can be identified that are also seen in the field of DHGR with SML and in the FSL domain \citep{hazraEtAl2019, fanEtAl2020, zengEtal2021, mauroEtal2022, shenEtal2022, rzeckiKrzysztof2020, rahimianEtal2021, zabihiEtal2022, luiMan2012, wanEtal2013, jiangEtal2013, cabreraEtal2017, jiangEtal2017, zhangEtal2017, chandrasekharMhala2019, luEtal2019, maEtal2020, wuEtal2021}. Image-based research from the past decade, primarily driven by the FSL-ChaLearn Challenge with a broad dataset of over 200 gestures \citep{WanEtal2016}, has pursued data augmentation methods exclusively, using domain expert knowledge to enrich the data. More recent research based on non-image data has moved away from this approach since it relies heavily on the existence of domain experts to design the data augmentation rules. Instead, algorithm and model methods, especially embedding learning models have been applied with great success \citep{hazraEtAl2019, fanEtAl2020, zengEtal2021, mauroEtal2022, shenEtal2022,rzeckiKrzysztof2020, zabihiEtal2022}. This trend is yet to be observed in image-based publications, as only two studies decided to use algorithm and model methods \citep{maEtal2020, wuEtal2021}. Despite achieving great accuracy, both models are limited by the fact that they rely on RGB-D data and are trained on datasets with a low number of classes, compared to FSL research in image, object, or text classification.

\section{Methodology}\label{sec3}
\subsection{Experimental setup}\label{sec3}

In order to investigate the feasibility of RGB image-based FSL models for DHGR, we trained and evaluated six models with differing task configurations, defined by K-Shot and N-Way. For K-Shot, we chose between one, two, and five, whereas five and ten were chosen as options for N-Way. To quantify the additional value of FSL models relative to SML models, we further trained SML models with an increasing amount of training samples to measure how many examples the standard approach requires to exceed the accuracy of our FSL models. 

\subsection{Data}\label{sec3}

For training and evaluation of our models, we used the Jester dataset \citep{MaterzynskaEtal2019}, a set of 148,092 videos that features 27 different dynamic hand gestures, performed by 1,376 actors. The dataset was selected from a large number of potential datasets through a review process because it shows the greatest diversity among the RGB-only datasets. All classes except "Doing other things" have been used since that class contained too many unlabeled gestures, which is not expedient for our purpose. Aiming to lower the complexity of the models' tasks that comes with processing raw pixel information, we transformed the image sequences into sequences of 21 skeletal points in three dimensions using Google's library MediaPipe Hands \citep{zhangEtal2020}. 

\begin{figure}[h]
\centering
\includegraphics[width=0.7\textwidth]{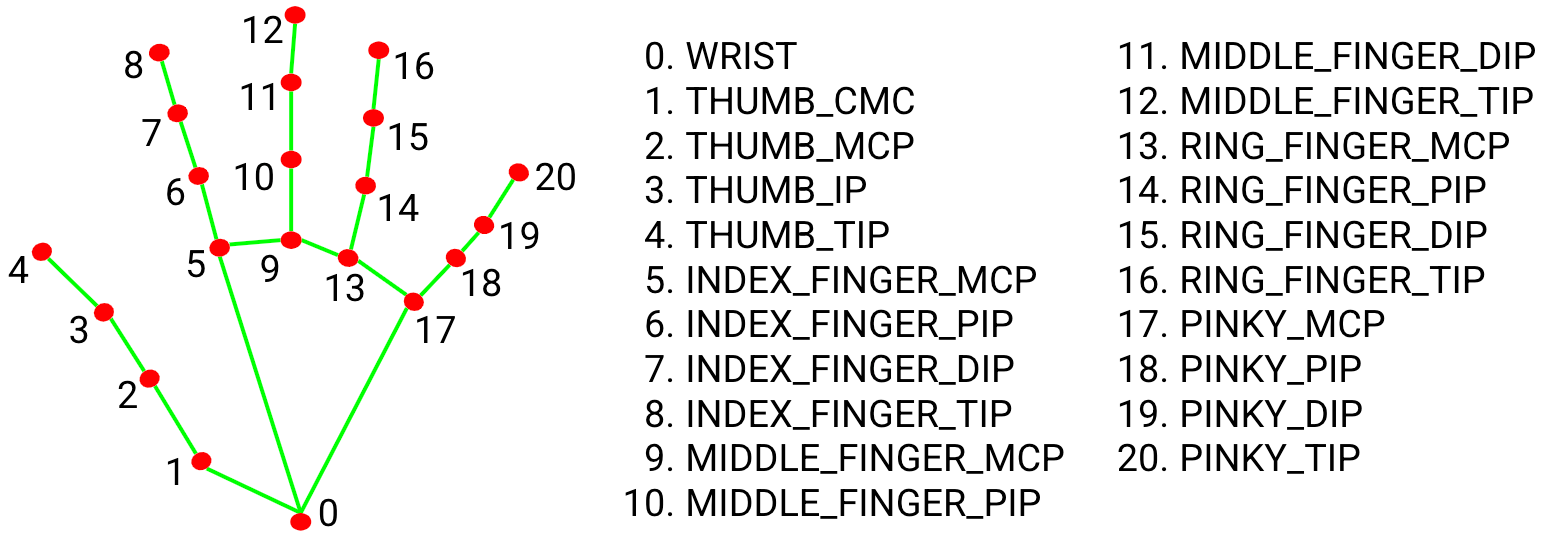}
\caption{Hand reference points obtained by MediaPipe Hands in three dimensions for each hand detected within an image \citep{landmarks_image}.}\label{fig:hand_landmarks}
\end{figure}

As the original dataset holds just 27 different dynamic hand gestures, which contradicts our goal of using a broad set of classes, we combined the existing classes creating a larger number of more complex gestures. For this purpose, we combined 250 samples for each class by appending samples from another class. The resulting "jump" of the hand in the three-dimensional space at the point where classes have been appended was adjusted accordingly. In addition to increasing the variety of classes, this step also made each class more versatile and complicated to recognize, which benefited the robustness of our models.

\begin{figure}[h]
\centering
\includegraphics[width=1\textwidth]{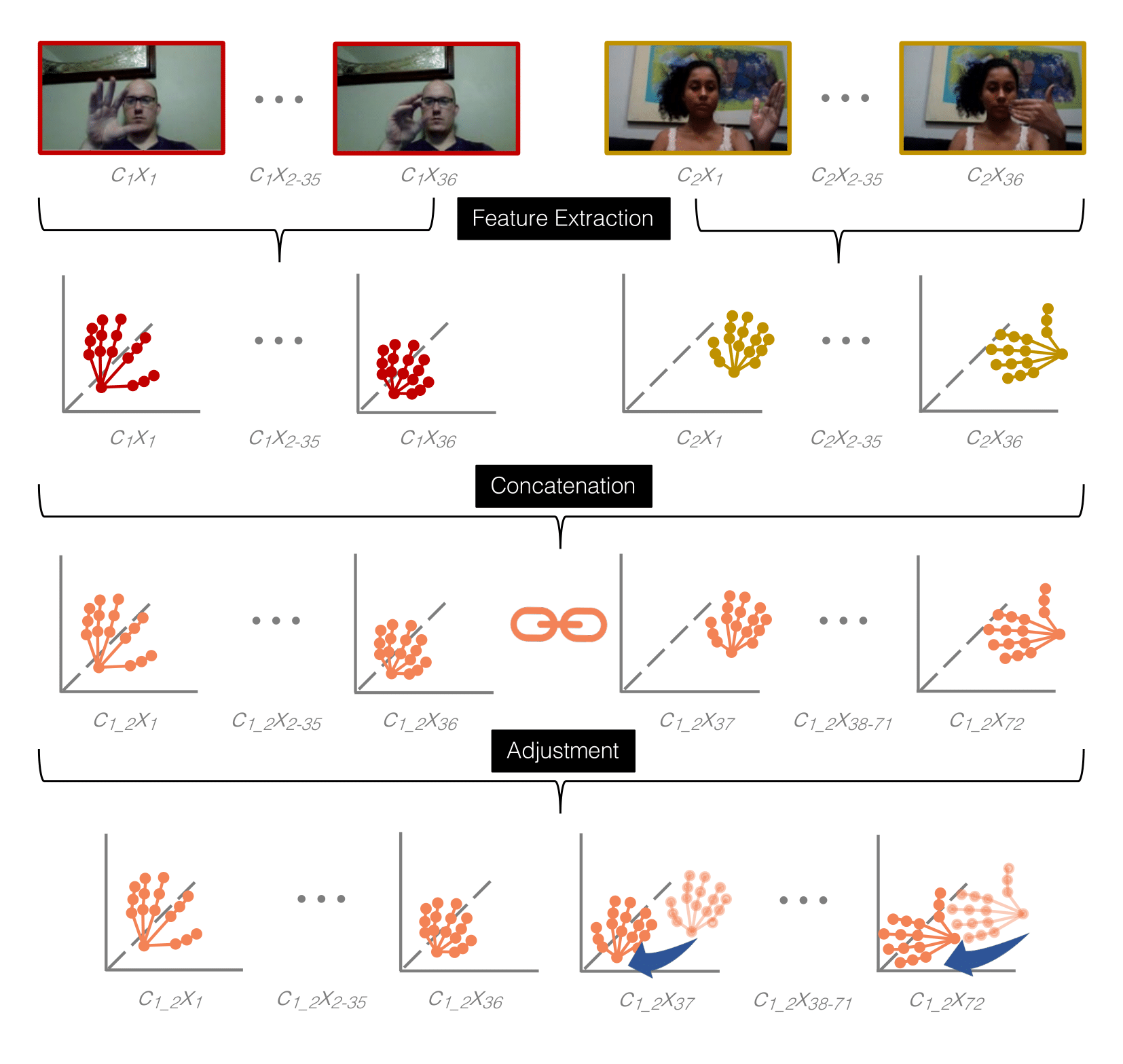}
\caption{Process of our data preparation that turned the original image-based Jester dataset into a set of skeletal points with a multitude of classes.}\label{fig:data_pipeline}
\end{figure}

The result of our data treatment process is a dataset with 169,000 samples, containing 676 different classes. Each class combines two original dynamical hand gestures from the Jester dataset with 250 samples. Every sample comprises a sequence of 72 steps, whereby the active hand gesture fills 49 steps on average. 36 percent of all sequence steps are empty since no hand was found in the corresponding image, as indicated by MediaPipe. Each sequence step holds 63 values assembled from 21 hand reference points with x, y, and z coordinates representing the hand skeleton in a three-dimensional space. For training, tuning, and evaluation, the dataset was split by classes, whereas care was taken that no original classes appeared in more than one sub-dataset. By taking this step, we avoided target leakage, i.e. enabling the model to see parts of the data in training and testing, but instead evaluated completely unknown data exclusively. As a result, we trained the models on a training set of 133 classes with 33,250 samples, tuned hyper-parameters on a set of 57 unseen classes with 14,250 samples, and tested the models on 57 unseen classes with 14,250 samples.

\subsection{FSL model design}\label{sec3}

Our architecture builds upon the Relation Network by \cite{sungEtal2018}, which originates in image classification, but has proven to work well when dealing with sequential data, compared to other embedding learners \citep{gengEtal2019, suiEtal2020}. The Relation Network is designed to consist of two modules in which data is passed through an embedding module and a relation module. Samples in the query set and samples in the support set are fed through the embedding module, which calculates feature maps for both sets and summarizes the support set feature maps if K-Shot is greater than one. All feature maps are concatenated afterwards. The combined feature maps are then transmitted to the relation module, which creates a relation score, a scalar value in the range of zero to one, representing the degree of similarity between each instance of the support set and the query set. Despite being used as a classification algorithm, Relation Networks train to minimize the RMSE, calculated using the relation scores. The Adam optimizer was utilized to minimize this loss. 

\begin{figure}[h]
\centering
\includegraphics[width=1\textwidth]{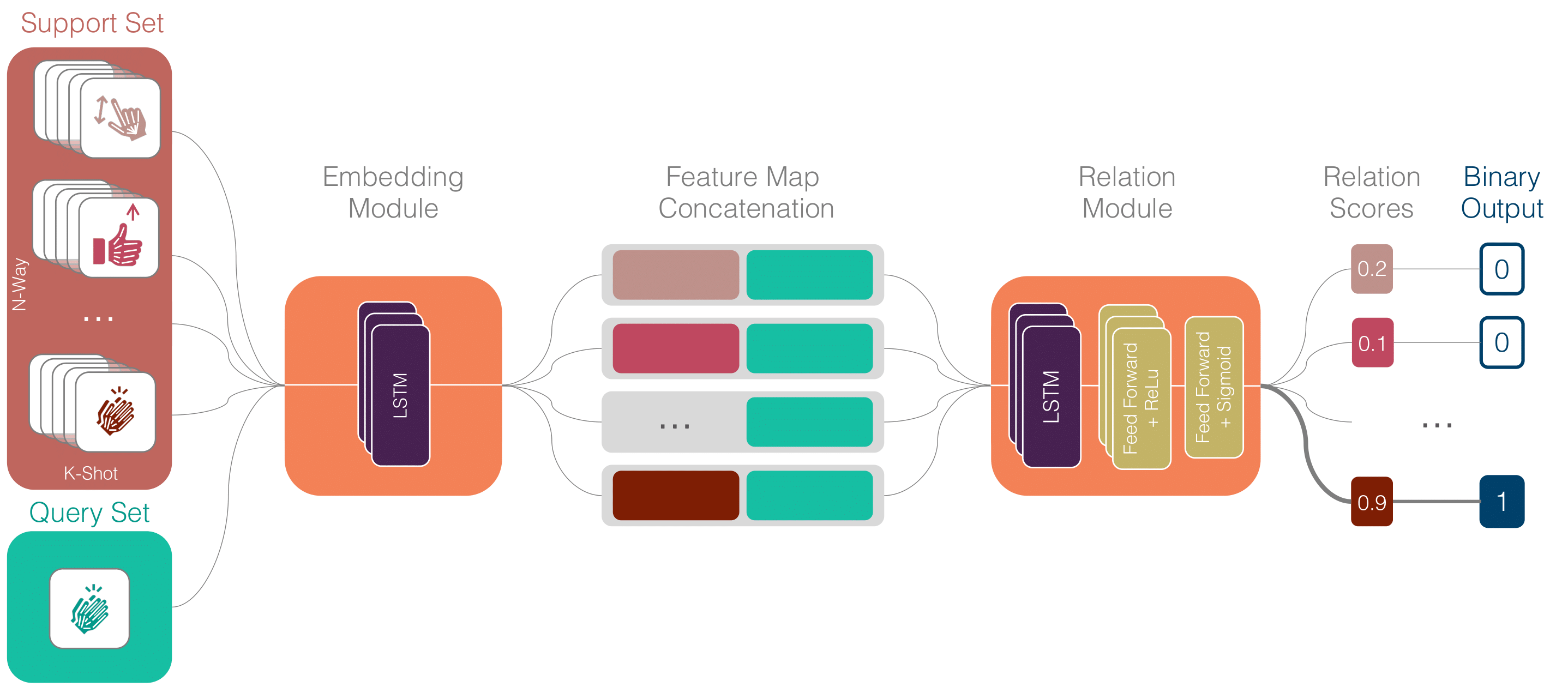}
\caption{Unlike the original Relation Network, we used LSTM cells instead of Convolution Layer in the backbone modules to deal with the sequential dependencies in our data. Inside the Relation Module, the data is parsed by Feed Forward (FF) layers with ReLu activation functions and then converted to Relation Scores using a sigmoid function.}\label{fig:model_architecture}
\end{figure}

The number of training episodes, learning rates, number and sizes of LSTM cells and FF layers have been selected in a hyper-parameter tuning within each FSL task configuration, whereas the goal was to minimize the RMSE with the lowest model complexity. Ranges of potential parameter values and final configurations for each model may be found in our GitHub repository \citep{schluesener}.

\subsection{Reference model design}\label{sec3}

Models used as a reference, intending to calculate the additional value of FSL models relative to SML models, were designed and trained to have equal prerequisites as the backbones of the FSL models. This ensured that the comparison of both approaches was as fair as possible. Input data got passed through one LSTM cell of size 64, one FF layer of size 256 with ReLu activation function, and a final Softmax function of size K-Shot to produce probabilities. Every model was trained on N-Way classes of our dataset, whose class-individual  accuracy of the FSL model evaluation were close to the general FSL models' accuracy. For training, we provided the models with an increasing number of samples per class, starting from one, along the powers of two series, up to 512. The additional value was then calculated by subtracting the number of samples per class our reference models needed in order to outperform our FSL models, measured by accuracy, from K-Shot of the outperformed FSL model.

\section{Results and Discussion}\label{sec4}
\subsection{Performance}\label{sec4}

Evaluated through 1000 test runs on our individual test dataset with 57 unseen classes, our models achieved an average accuracy ranging from 77,32\% to 88,84\% on the 5-Way task and from 70,37\% to 81,28\% on the 10-Way task. Comparing these results to existing FSL models is challenging due to differences in the underlying data, for which we focused on RGB-based data exclusively and used significantly more classes with higher complexities for training, tuning, and testing, making it a more challenging task than equivalent FSL-research approaches. This is why we have also tested our models on 15 classes of the original Jester dataset, those not seen by the algorithm during training, to provide a reasonable benchmark for the model. Within this comparison, the proposed models occupy mid-table positions in the ranking of the Jester paper, with an accuracy ranging from 83.13\% to 91.48\% on unseen data, supported by only 1 to 5 samples per class. When making this comparison, it should be noted that accuracy as a performance indicator can be biased by low-class tasks, as shown by the difference between our results at N-Way = 5 and N-Way = 10. Additionally, it should be considered that most of the models on the Jester leaderboard work independently with raw pixel data, while our model relies on MediaPipe's ability to provide high-quality skeleton data. Nevertheless, the benchmark indicates that our models achieve competitive results with a significantly smaller amount of data.

\begin{table}[h]
	\caption{Overview on our models' performance}
	\centering
            \begin{tabular}{@{}lcc@{}}
                \toprule
                Model & Modified Jester & Jester\\
                \midrule
                5-Way 1-Shot & 77,32\% & 87,03\%\\
                5-Way 2-Shot & 84,01\% & 89,28\%\\
                5-Way 5-Shot & 88,84\% & 91,48\%\\
                10-Way 1-Shot & 70,37\% & 83,13\%\\
                10-Way 2-Shot & 75,71\% & 85,59\%\\
                10-Way 5-Shot & 81,28\% & 89,87\%\\
		\bottomrule
	\end{tabular}
	\label{tab:table}
\end{table}

\subsection{Benefit}\label{sec4}

Evidence supporting the claim that adapting FSL models to new tasks takes less data than SML models was given by our comparison experiment. With total savings, measured in sample data points, required less when using our FSL models, of 315 up to 630 for five gestures and 630 to 1230 for ten gestures to be recognized, we quantify a definite advantage of the FSL approach. Thereby we can observe that the total saving increases strongly with the number of the task's classes, and along with it, time and costs that can be spared. However, we also find that the SML models can outperform our FSL models, which makes them superior to our approach when accuracy is the primary concern. Furthermore, we have to limit the transferability of our findings on the overall area of FSL research, since we provided our models with highly processed skeleton data instead of raw pixel or raw text information, which is mostly used in FSL. 

\begin{table}[h]
	\caption{Overview of our models' savings}
	\centering
            \begin{tabular}{lccccc}
            \toprule
            Task & Samples\footnotemark[1] FSL & Accuracy FSL & Samples\footnotemark[1] SML & Accuracy SML & Savings\\
            \midrule
            5-Way  & 1 & 77,32\% & 64 & 81,40\% & 315 \\
            5-Way  & 2 & 84,01\% & 128 & 92,31\% & 630 \\
            5-Way  & 5 & 88,84\% & 128 & 92,31\% & 615 \\
            10-Way  & 1 & 70,37\% & 64 & 76,33\% & 630 \\
            10-Way  & 2 & 75,71\% & 128 & 81,67\% & 1260 \\
            10-Way  & 5 & 81,28\% & 128 & 81,67\% & 1230 \\
		\bottomrule
            \multicolumn{6}{l}{Listed SML models were the first in the series of trained SML models to exceed the}\\
            \multicolumn{6}{l}{accuracy of our FSL models.}\\
	\end{tabular}
	\label{tab:table}
\end{table}
\footnotetext[1]{provided per class for training.}

\section{Conclusion}\label{sec5}

In this paper, we present six FSL models, trained for quick adaptation on recognizing five or ten dynamic hand gestures, given one, two, or five samples per gesture. All models work with skeleton data, extracted from RGB videos with Google's MediaPipe library, which is fed through a Relation Network FSL architecture with LSTM and FF Layers as a backbone. Thus, we prove that RGB-D data is not mandatory to recognize dynamic hand gestures from image data. Additionally, we provide evidence for the claim that FSL saves time and resources over SML models, by quantifying the savings to be up to 1230 samples when using our models instead of training a SML model. Besides these findings, our work identifies research gaps, such as the lack of an RGB-based dataset for DHGR with a large number of distinct gestures. Future work might also advance our models to work without extraction of hand skeleton points, or to be designed as a zero-shot architecture allowing new, unknown gestures to be added even faster.

\bibliographystyle{plainnat}
\bibliography{arxiv-ref}

\begin{thebibliography}{51}
\providecommand{\natexlab}[1]{#1}
\providecommand{\url}[1]{\texttt{#1}}
\expandafter\ifx\csname urlstyle\endcsname\relax
  \providecommand{\doi}[1]{doi: #1}\else
  \providecommand{\doi}{doi: \begingroup \urlstyle{rm}\Url}\fi

\bibitem[Adeyanju et~al.(2021)Adeyanju, Bello, and Adegboye]{adeyanjuEtal2021}
IA~Adeyanju, OO~Bello, and MA~Adegboye.
\newblock Machine learning methods for sign language recognition: A critical
  review and analysis.
\newblock \emph{Intelligent Systems with Applications}, 12, 2021.

\bibitem[Berndt and Clifford(1994)]{berndtClifford1994}
Donald~J Berndt and James Clifford.
\newblock Using dynamic time warping to find patterns in time series.
\newblock In \emph{KDD workshop}, volume~10, pages 359--370. Seattle, WA, USA:,
  1994.

\bibitem[Cabrera and Wachs(2017)]{cabreraEtal2017}
Maria~Eugenia Cabrera and Juan~Pablo Wachs.
\newblock A human-centered approach to one-shot gesture learning.
\newblock \emph{Frontiers in Robotics and AI}, 4:\penalty0 8, 2017.

\bibitem[Chandrasekhar and Mhala(2019)]{chandrasekharMhala2019}
S~Chandrasekhar and NN~Mhala.
\newblock Improving the robustness of hand gesture recognition with
  one-shot-learning features of static hog and svm by using kinect v2 method.
\newblock In \emph{2019 International Conference on Intelligent Sustainable
  Systems (ICISS)}, pages 15--19. IEEE, 2019.

\bibitem[Chen et~al.(2019)Chen, Liu, Kira, Wang, and Huang]{chenEtal2019}
Wei-Yu Chen, Yen-Cheng Liu, Zsolt Kira, Yu-Chiang~Frank Wang, and Jia-Bin
  Huang.
\newblock A closer look at few-shot classification.
\newblock In \emph{International Conference on Learning Representations}, 2019.
\newblock URL \url{https://openreview.net/forum?id=HkxLXnAcFQ}.

\bibitem[Deng et~al.(2009)Deng, Dong, Socher, Li, Li, and
  Fei-Fei]{dengEtal2009}
Jia Deng, Wei Dong, Richard Socher, Li-Jia Li, Kai Li, and Li~Fei-Fei.
\newblock Imagenet: A large-scale hierarchical image database.
\newblock In \emph{2009 IEEE conference on computer vision and pattern
  recognition}, pages 248--255. Ieee, 2009.

\bibitem[Dhillon et~al.(2019)Dhillon, Chaudhari, Ravichandran, and
  Soatto]{dhillonEtal2019}
Guneet~S Dhillon, Pratik Chaudhari, Avinash Ravichandran, and Stefano Soatto.
\newblock A baseline for few-shot image classification.
\newblock \emph{arXiv preprint arXiv:1909.02729}, 2019.

\bibitem[Fan et~al.(2020{\natexlab{a}})Fan, Zhuo, Tang, and Tai]{fanEtal2020b}
Qi~Fan, Wei Zhuo, Chi-Keung Tang, and Yu-Wing Tai.
\newblock Few-shot object detection with attention-rpn and multi-relation
  detector.
\newblock In \emph{Proceedings of the IEEE/CVF Conference on Computer Vision
  and Pattern Recognition}, pages 4013--4022, 2020{\natexlab{a}}.

\bibitem[Fan et~al.(2020{\natexlab{b}})Fan, Zheng, and Feng]{fanEtAl2020}
Zhongyu Fan, Haifeng Zheng, and Xinxin Feng.
\newblock A meta-learning-based approach for hand gesture recognition using
  fmcw radar.
\newblock In \emph{2020 International Conference on Wireless Communications and
  Signal Processing (WCSP)}, pages 522--527. IEEE, 2020{\natexlab{b}}.

\bibitem[Gallo et~al.(2011)Gallo, Placitelli, and Ciampi]{galloEtal2011}
Luigi Gallo, Alessio~Pierluigi Placitelli, and Mario Ciampi.
\newblock Controller-free exploration of medical image data: Experiencing the
  kinect.
\newblock In \emph{2011 24th International Symposium on Computer-Based Medical
  Systems (CBMS)}, pages 1--6, 2011.
\newblock \doi{10.1109/CBMS.2011.5999138}.

\bibitem[Geng et~al.(2019)Geng, Li, Li, Zhu, Jian, and Sun]{gengEtal2019}
Ruiying Geng, Binhua Li, Yongbin Li, Xiaodan Zhu, Ping Jian, and Jian Sun.
\newblock Induction networks for few-shot text classification.
\newblock \emph{arXiv preprint arXiv:1902.10482}, 2019.

\bibitem[{Google LLC}(2020)]{landmarks_image}
{Google LLC}.
\newblock Mediapipe hands, 2020.
\newblock URL \url{https://google.github.io/mediapipe/solutions/hands.html}.
\newblock Accessed: 2022-12-11.

\bibitem[Hazra and Santra(2019)]{hazraEtAl2019}
Souvik Hazra and Avik Santra.
\newblock Short-range radar-based gesture recognition system using 3d cnn with
  triplet loss.
\newblock \emph{IEEE Access}, 7:\penalty0 125623--125633, 2019.

\bibitem[Hochreiter and Schmidhuber(1997)]{hochreiterSchmidhuber1997}
Sepp Hochreiter and Jürgen Schmidhuber.
\newblock Long short-term memory.
\newblock \emph{Neural computation}, 9:\penalty0 1735--1780, 12 1997.
\newblock \doi{10.1162/neco.1997.9.8.1735}.

\bibitem[Jiang et~al.(2017)Jiang, Ren, Lee, Shi, Liu, and Zhao]{jiangEtal2017}
Feng Jiang, Jie Ren, Changhoon Lee, Wuzhen Shi, Shaohui Liu, and Debin Zhao.
\newblock Spatial and temporal pyramid-based real-time gesture recognition.
\newblock \emph{Journal of Real-Time Image Processing}, 13\penalty0
  (3):\penalty0 599--611, 2017.

\bibitem[Jiang et~al.(2013)Jiang, Duerstock, and Wachs]{jiangEtal2013}
Hairong Jiang, Bradley~S Duerstock, and Juan~P Wachs.
\newblock A machine vision-based gestural interface for people with upper
  extremity physical impairments.
\newblock \emph{IEEE Transactions on Systems, Man, and Cybernetics: Systems},
  44\penalty0 (5):\penalty0 630--641, 2013.

\bibitem[Kang et~al.(2019)Kang, Liu, Wang, Yu, Feng, and Darrell]{kangEtal2019}
Bingyi Kang, Zhuang Liu, Xin Wang, Fisher Yu, Jiashi Feng, and Trevor Darrell.
\newblock Few-shot object detection via feature reweighting.
\newblock In \emph{Proceedings of the IEEE/CVF International Conference on
  Computer Vision}, pages 8420--8429, 2019.

\bibitem[Lake et~al.(2019)Lake, Salakhutdinov, and Tenenbaum]{lakeEtal2019}
Brenden~M Lake, Ruslan Salakhutdinov, and Joshua~B Tenenbaum.
\newblock The omniglot challenge: a 3-year progress report.
\newblock \emph{Current Opinion in Behavioral Sciences}, 29:\penalty0 97--104,
  2019.

\bibitem[LeCun et~al.(2015)LeCun, Bengio, and Hinton]{lecunEtal2015}
Yann LeCun, Yoshua Bengio, and Geoffrey Hinton.
\newblock Deep learning.
\newblock \emph{nature}, 521\penalty0 (7553):\penalty0 436--444, 2015.

\bibitem[Lee and Hong(2010)]{leeHong2010}
Doe-Hyung Lee and Kwang-Seok Hong.
\newblock Game interface using hand gesture recognition.
\newblock In \emph{5th International Conference on Computer Sciences and
  Convergence Information Technology}, pages 1092--1097, 2010.
\newblock \doi{10.1109/ICCIT.2010.5711226}.

\bibitem[Liu et~al.(2018)Liu, Yu, Yu, Zhang, Wan, and Wang]{liuEtal2018}
Bing Liu, Xuchu Yu, Anzhu Yu, Pengqiang Zhang, Gang Wan, and Ruirui Wang.
\newblock Deep few-shot learning for hyperspectral image classification.
\newblock \emph{IEEE Transactions on Geoscience and Remote Sensing},
  57\penalty0 (4):\penalty0 2290--2304, 2018.

\bibitem[Lu et~al.(2019)Lu, Qin, Li, Zhang, Xu, and Hu]{luEtal2019}
Zhi Lu, Shiyin Qin, Lianwei Li, Dinghao Zhang, Kuanhong Xu, and Zhongying Hu.
\newblock One-shot learning hand gesture recognition based on lightweight 3d
  convolutional neural networks for portable applications on mobile systems.
\newblock \emph{IEEE Access}, 7:\penalty0 131732--131748, 2019.

\bibitem[Lui(2012)]{luiMan2012}
Yui~Man Lui.
\newblock Human gesture recognition on product manifolds.
\newblock \emph{The Journal of Machine Learning Research}, 13\penalty0
  (1):\penalty0 3297--3321, 2012.

\bibitem[Ma et~al.(2020)Ma, Zhang, Wang, Qi, and Chen]{maEtal2020}
Chunyong Ma, Shengsheng Zhang, Anni Wang, Yongyang Qi, and Ge~Chen.
\newblock Skeleton-based dynamic hand gesture recognition using an enhanced
  network with one-shot learning.
\newblock \emph{Applied Sciences}, 10\penalty0 (11), 2020.

\bibitem[Materzynska et~al.(2019)Materzynska, Berger, Bax, and
  Memisevic]{MaterzynskaEtal2019}
Joanna Materzynska, Guillaume Berger, Ingo Bax, and Roland Memisevic.
\newblock The jester dataset: A large-scale video dataset of human gestures.
\newblock In \emph{Proceedings of the IEEE/CVF International Conference on
  Computer Vision Workshops}, pages 0--0, 2019.

\bibitem[Mauro et~al.(2022)Mauro, Chmurski, Servadei, Pegalajar-Cuellar, and
  Morales-Santos]{mauroEtal2022}
Gianfranco Mauro, Mateusz Chmurski, Lorenzo Servadei, Manuel Pegalajar-Cuellar,
  and Diego~P Morales-Santos.
\newblock Few-shot user-definable radar-based hand gesture recognition at the
  edge.
\newblock \emph{IEEE Access}, 10:\penalty0 29741--29759, 2022.

\bibitem[Mitchell(1997)]{mitchell1997}
Tom~M. Mitchell.
\newblock \emph{Machine learning, International Edition}.
\newblock McGraw-Hill Series in Computer Science. McGraw-Hill, 1997.
\newblock ISBN 978-0-07-042807-2.
\newblock URL \url{https://www.worldcat.org/oclc/61321007}.

\bibitem[Noraini et~al.(2021)Noraini, Mumtaz~Begum, and
  Nazean]{norainiEtal2021}
Mohamed Noraini, Mustafa Mumtaz~Begum, and Jomhari Nazean.
\newblock A review of the hand gesture recognition system: Current progress and
  future directions.
\newblock \emph{IEEE Access}, 2021.

\bibitem[Oudah et~al.(2020)Oudah, Al-Naji, and Chahl]{oudahEtal2020}
Munir Oudah, Ali Al-Naji, and Javaan Chahl.
\newblock Hand gesture recognition based on computer vision: a review of
  techniques.
\newblock \emph{Journal of Imaging}, 6\penalty0 (8):\penalty0 73, 2020.

\bibitem[Pisharady and Saerbeck(2015)]{pisharadyEtal2015}
Pramod~Kumar Pisharady and Martin Saerbeck.
\newblock Recent methods and databases in vision-based hand gesture
  recognition: A review.
\newblock \emph{Computer Vision and Image Understanding}, 141:\penalty0
  152--165, 2015.

\bibitem[Rabiner and Juang(1986)]{rabinderJuang1986}
L.~Rabiner and B.~Juang.
\newblock An introduction to hidden markov models.
\newblock \emph{IEEE ASSP Magazine}, 3\penalty0 (1):\penalty0 4--16, 1986.
\newblock \doi{10.1109/MASSP.1986.1165342}.

\bibitem[Rahimian et~al.(2021)Rahimian, Zabihi, Asif, Farina, Atashzar, and
  Mohammadi]{rahimianEtal2021}
Elahe Rahimian, Soheil Zabihi, Amir Asif, Dario Farina, Seyed~Farokh Atashzar,
  and Arash Mohammadi.
\newblock Few-shot learning for decoding surface electromyography for hand
  gesture recognition.
\newblock In \emph{ICASSP 2021-2021 IEEE International Conference on Acoustics,
  Speech and Signal Processing (ICASSP)}, pages 1300--1304. IEEE, 2021.

\bibitem[Rzecki(2020)]{rzeckiKrzysztof2020}
Krzysztof Rzecki.
\newblock Classification algorithm for person identification and gesture
  recognition based on hand gestures with small training sets.
\newblock \emph{Sensors}, 20\penalty0 (24):\penalty0 7279, 2020.

\bibitem[Schlüsener(2022)]{schluesener}
Niels Schlüsener.
\newblock S-strhange, 2022.
\newblock URL
  \url{https://github.com/nielsschluesener/Fast-Learning-Hand-Gesture-Recognition}.
\newblock Accessed: 2022-10-24.

\bibitem[Shen et~al.(2022)Shen, Zheng, Feng, and Hu]{shenEtal2022}
Xiangyu Shen, Haifeng Zheng, Xinxin Feng, and Jinsong Hu.
\newblock Ml-hgr-net: A meta-learning network for fmcw radar based hand gesture
  recognition.
\newblock \emph{IEEE Sensors Journal}, 2022.

\bibitem[Snell et~al.(2017)Snell, Swersky, and Zemel]{snellEtal2017}
Jake Snell, Kevin Swersky, and Richard Zemel.
\newblock Prototypical networks for few-shot learning.
\newblock \emph{Advances in neural information processing systems}, 30, 2017.

\bibitem[Sui et~al.(2020)Sui, Chen, Mao, Qiu, Liu, and Zhao]{suiEtal2020}
Dianbo Sui, Yubo Chen, Binjie Mao, Delai Qiu, Kang Liu, and Jun Zhao.
\newblock Knowledge guided metric learning for few-shot text classification.
\newblock \emph{arXiv preprint arXiv:2004.01907}, 2020.

\bibitem[Sung et~al.(2018)Sung, Yang, Zhang, Xiang, Torr, and
  Hospedales]{sungEtal2018}
Flood Sung, Yongxin Yang, Li~Zhang, Tao Xiang, Philip Torr, and Timothy
  Hospedales.
\newblock Learning to compare: Relation network for few-shot learning.
\newblock pages 1199--1208, 06 2018.
\newblock \doi{10.1109/CVPR.2018.00131}.

\bibitem[Vinyals et~al.(2016)Vinyals, Blundell, Lillicrap, Wierstra,
  et~al.]{vinyalsEtal2016}
Oriol Vinyals, Charles Blundell, Timothy Lillicrap, Daan Wierstra, et~al.
\newblock Matching networks for one shot learning.
\newblock \emph{Advances in neural information processing systems}, 29, 2016.

\bibitem[Wan et~al.(2013)Wan, Ruan, Li, and Deng]{wanEtal2013}
Jun Wan, Qiuqi Ruan, Wei Li, and Shuang Deng.
\newblock One-shot learning gesture recognition from rgb-d data using bag of
  features.
\newblock \emph{The Journal of Machine Learning Research}, 14\penalty0
  (1):\penalty0 2549--2582, 2013.

\bibitem[Wan et~al.(2016)Wan, Zhao, Zhou, Guyon, Escalera, and Li]{WanEtal2016}
Jun Wan, Yibing Zhao, Shuai Zhou, Isabelle Guyon, Sergio Escalera, and Stan~Z.
  Li.
\newblock Chalearn looking at people rgb-d isolated and continuous datasets for
  gesture recognition.
\newblock In \emph{Proceedings of the IEEE Conference on Computer Vision and
  Pattern Recognition (CVPR) Workshops}, June 2016.

\bibitem[Wang et~al.(2020)Wang, Yao, Kwok, and Ni]{wangEtal2020}
Wang, Quanming Yao, James~T. Kwok, and Lionel~M. Ni.
\newblock Generalizing from a few examples: A survey on few-shot learning.
\newblock \emph{ACM Comput. Surv.}, 53\penalty0 (3), 2020.
\newblock ISSN 0360-0300.
\newblock \doi{10.1145/3386252}.
\newblock URL \url{https://doi.org/10.1145/3386252}.

\bibitem[Wu et~al.(2021)Wu, Zhang, and Zhao]{wuEtal2021}
Jinting Wu, Yujia Zhang, and Xiaoguang Zhao.
\newblock A prototype-based generalized zero-shot learning framework for hand
  gesture recognition.
\newblock In \emph{2020 25th International Conference on Pattern Recognition
  (ICPR)}, pages 3435--3442. IEEE, 2021.

\bibitem[Yan et~al.(2018)Yan, Zheng, and Cao]{yanEtal2018}
Leiming Yan, Yuhui Zheng, and Jie Cao.
\newblock Few-shot learning for short text classification.
\newblock \emph{Multimedia Tools and Applications}, 77\penalty0 (22):\penalty0
  29799--29810, 2018.

\bibitem[Yang et~al.(2017)Yang, Pan, and Li]{yangEtal2017}
Jinxing Yang, Jianhong Pan, and Jun Li.
\newblock semg-based continuous hand gesture recognition using gmm-hmm and
  threshold model.
\newblock In \emph{2017 IEEE International Conference on Robotics and
  Biomimetics (ROBIO)}, pages 1509--1514, 2017.
\newblock \doi{10.1109/ROBIO.2017.8324631}.

\bibitem[Yasen and Jusoh(2019)]{yasenJusoh2019}
Mais Yasen and Shaidah Jusoh.
\newblock A systematic review on hand gesture recognition techniques,
  challenges and applications.
\newblock \emph{PeerJ Computer Science}, 5:\penalty0 e218, 2019.

\bibitem[Zabihi et~al.(2022)Zabihi, Rahimian, Asif, and
  Mohammadi]{zabihiEtal2022}
Soheil Zabihi, Elahe Rahimian, Amir Asif, and Arash Mohammadi.
\newblock Trahgr: Few-shot learning for hand gesture recognition via
  electromyography.
\newblock \emph{arXiv preprint arXiv:2203.16336}, 2022.

\bibitem[Zeng et~al.(2021)Zeng, Wu, and Ye]{zengEtal2021}
Xianglong Zeng, Chaoyang Wu, and Wen-Bin Ye.
\newblock User-definable dynamic hand gesture recognition based on doppler
  radar and few-shot learning.
\newblock \emph{IEEE Sensors Journal}, 21\penalty0 (20):\penalty0 23224--23233,
  2021.
\newblock \doi{10.1109/JSEN.2021.3107943}.

\bibitem[Zhang et~al.(2020)Zhang, Bazarevsky, Vakunov, Tkachenka, Sung, Chang,
  and Grundmann]{zhangEtal2020}
Fan Zhang, Valentin Bazarevsky, Andrey Vakunov, Andrei Tkachenka, George Sung,
  Chuo-Ling Chang, and Matthias Grundmann.
\newblock Mediapipe hands: On-device real-time hand tracking.
\newblock \emph{arXiv preprint arXiv:2006.10214}, 2020.

\bibitem[Zhang et~al.(2017)Zhang, Zhang, Jiang, Qi, Zhang, Guo, and
  Zhou]{zhangEtal2017}
Lei Zhang, Shengping Zhang, Feng Jiang, Yuankai Qi, Jun Zhang, Yuliang Guo, and
  Huiyu Zhou.
\newblock Bomw: Bag of manifold words for one-shot learning gesture recognition
  from kinect.
\newblock \emph{IEEE Transactions on Circuits and Systems for Video
  Technology}, 28\penalty0 (10):\penalty0 2562--2573, 2017.

\bibitem[Zhang(2012)]{zhang2012}
Zhengyou Zhang.
\newblock Microsoft kinect sensor and its effect.
\newblock \emph{IEEE multimedia}, 19\penalty0 (2):\penalty0 4--10, 2012.

\end{thebibliography}

\end{document}